\documentclass[]{interact}

\usepackage{epstopdf}
\usepackage[caption=false]{subfig}
\usepackage{url}

\usepackage[doublespacing]{setspace}
\usepackage{algorithm,algpseudocode}
\usepackage{multirow}
\usepackage[numbers,sort&compress]{natbib}
\setcitestyle{authoryear,open={(},close={)}}
\renewcommand\bibfont{\fontsize{10}{12}\selectfont}
\makeatletter
\def\NAT@def@citea{\def\@citea{\NAT@separator}}
\makeatother

\theoremstyle{plain}
\newtheorem{theorem}{Theorem}[section]
\newtheorem{lemma}[theorem]{Lemma}

\theoremstyle{definition}
\newtheorem{definition}[theorem]{Definition}

\theoremstyle{remark}

\algnewcommand{\Inputs}[1]{%
  \State \textbf{Inputs:}
  \Statex \hspace*{\algorithmicindent}\parbox[t]{.8\linewidth}{\raggedright #1}
}
\algnewcommand{\Initialize}[1]{%
  \State \textbf{Initialize:}
  \Statex \hspace*{\algorithmicindent}\parbox[t]{.8\linewidth}{\raggedright #1}
}
\usepackage{soul}
\usepackage{color}

\begin{document}


\title{Tensor decomposition to Compress Convolutional Layers in Deep Learning}
\author{
\name{$\text{Yinan Wang}^{1}$, $\text{Weihong ``Grace" Guo}^{2}$, $\text{and Xiaowei Yue}^{1*}$\thanks{$^{*} \text{corresponding author: Dr. Xiaowei Yue. Email: xwy@vt.edu}$}}
\affil{$^{1}$\text{Grado Department of Industrial and Systems Engineering,} \\ \text{\;\;Virginia Polytechnic Institute and State University, Blacksburg, VA, 24061}}
\affil{$^{2}$\text{Department of Industrial and Systems Engineering,} \\ \text{\;\;Rutgers University, New Brunswick, NJ, 08901}}
}
\maketitle

\begin{abstract}
Feature extraction for tensor data serves as an important step in many tasks such as anomaly detection, process monitoring, image classification, and quality control. Although many methods have been proposed for tensor feature extraction, there are still two challenges that need to be addressed: 1) how to reduce the computation cost for high dimensional and large volume tensor data; 2) how to interpret the output features and evaluate their significance. {The most recent methods in deep learning, such as Convolutional Neural Network (CNN), have shown outstanding performance in analyzing tensor data, but their wide adoption is still hindered by model complexity} and lack of interpretability. To fill this research gap, we propose to use CP-decomposition to approximately compress the convolutional layer (CPAC-Conv layer) in deep learning. The contributions of our work could be summarized into three aspects: (1) we adapt CP-decomposition to compress convolutional kernels and derive the expressions of both forward and backward propagations for our proposed CPAC-Conv layer; (2) compared with the original convolutional layer, the proposed CPAC-Conv layer can reduce the number of parameters without decaying prediction performance. It can combine with other layers to build novel deep Neural Networks; (3) the value of decomposed kernels indicates the significance of the corresponding feature map, {which provides us with insights to guide feature selection}.
\end{abstract}

\begin{keywords}
CP-decomposition, Tensor, Convolutional Neural Network (CNN), Model compression, Feature selection
\end{keywords}

\section{Introduction}
With the development of sensing technology in recent years, the high-rate and high-resolution image sensors have become ubiquitous in the smart manufacturing systems. Compared with other data types, i.e. spectrum, vector, etc., image data is more straightforward and easy to understand by human eyes. Images can convey rich and various information associated with 2D or 3D geometries, spatial-temporal structures, and multi-channel dynamic changes. Therefore, {they are becoming more and more critical in many industrial and system engineering applications, such as anomaly detection} (\cite{Hao17}), {spatiotemporal characterization} (\cite{shao2017dynamic}), {quality prediction} (\cite{Yifu20}), {high dimensional profile monitoring} (\cite{Sergin19}), {quality control and process optimization} (\cite{Chenang19, gao2020optimal}).

In the pipeline of image-based data analysis, feature extraction is one essential intermediate step {for both supervised and unsupervised learning}. Most of the applications are implemented based on specific features.  Deterministic and stochastic decompositions are effective ways to learn informative features (\cite{yue2019data}). For example, Principal component analysis
(PCA) and its variants are classical feature extraction techniques designed for learning a lower-dimensional representation of the original data (\cite{Nomikos94,Alex02}). {For tensor-format data, multi-linear PCA  and uncorrelated multi-linear PCA were proposed to find the orthogonal basis to capture the most of variation in the original multi-way tensor and avoid the correlations among features} (\cite{Lu08}). {Although the PCA and its variants can compress data and extract features in a simple and scalable way, their effectiveness is hindered by the assumption that the original data is the linear combination of principal components, and the interpretability of principal components. To specify the interpretable meaning of each decomposed component, Smooth-sparse Decomposition (SSD) was proposed to decompose an image into three components, which are smooth background, sparse anomalies, and random noise} (\cite{Hao17}).{ This decomposition achieved excellent performance in anomaly detection for images. Tensor decomposition techniques have also been developed for feature extraction. CANDECOMP/PARAFAC (CP) decomposition} (\cite{Kiers2000}) can approximate the original tensor with the summation of rank-one tensors, and Turker decomposition (\cite{Tuck1966c}) can decompose the original tensor into a core tensor along with multiple matrices. \cite{kolda2009tensor} and \cite{papalexakis2016tensors} did thorough literature reviews about tensor decomposition. \cite{yue2020tensor} proposed a tensor mixed effects model to decompose fixed effects and random effects in tensors and learn correlations along different dimensions. \cite{gao2020optimal} integrated tensor decomposition and ensemble learning by {utilizing mutual benefits} so as to improve quality evaluation. 

The aforementioned feature extraction techniques are mainly derived from the perspective of statistics. Some mathematical signal processing techniques (specifically computational harmonic analysis) can be used to extract features from image data. In the field of signal processing, transformation methods will transform an image into a new domain based on the spatial frequencies characteristics of the pixel intensity variations. Multi-way PCA (MPCA) and 2D Fast Fourier Transformation (FFT) have been applied to extract texture features from images (\cite{Paul92}). Wavelet transformation also showed its advantages in signal decomposition and has the potential to extend to multi-way tensor (\cite{Mallat89}). \cite{Jin99} firstly proposed wavelet-based profile monitoring for quality control. Wavelet-based feature extraction was further been adapted to profile data and multi-channel profile data for real-time detection and quality improvement in advanced manufacturing systems (\cite{paynabar2011characterization, Yue18}).

Recently developed deep learning methods have become another important branch in feature extraction techniques. Among all the structures of neural networks, Convolutional Neural Network (CNN) is designed to process multi-way tensor data and has shown its strength in many fields such as object detection, classification, segmentation, etc. ImageNet was the first CNN model proposed for classification tasks and outperformed the previous methods (\cite{Pereira12}). In the structure of ImageNet, the Convolutional layer (Conv layer) acts as the feature extractor and the Fully Connected layer (FC layer) acts as the classifier. Various works on designing novel CNN structures follow a similar idea, that is to design multiple Conv layers as the feature extractor, and use other types of layers according to the specific tasks (\cite{Simonyan15,  He16,  Girshick15}). {In the aforementioned models, the feature extractor, Conv layer, is working in the supervised approaches, which means all the features from images are extracted on the guidance of specific tasks.} {The Conv layer can also work as a feature extractor in the unsupervised approaches, for example, Auto-encoder is a family of unsupervised learning methods focusing on extracting features from multi-way tensors. Stacked Convolutional Auto-Encoders (CAE) was proposed for unsupervised hierarchical feature extraction } (\cite{Masci2}){, which uses the Conv layer as the building block}.

Although deep learning has shown its strength in {extracting informative features from} high dimensional and large-volume data {and applying these features for various tasks}, the model complexity and computational cost are still significant and it is hard to adapt the models to devices with limited computational resources. Additionally, large parameter size makes features from deep learning methods hard to interpret. {For example, in the metal casting process, the high-resolution image data ($512 \times 512$ pixels) is collected to detect the defects in the products. This problem can be defined as a classification task to distinguish normal and defect products. Although CNNs have excellent performance on image classification, model complexity and features interpretability are important factors hindered CNN's application in the manufacturing systems. These factors motivate us to explore the possibility to reduce the size of parameters and model complexity for better model generalization and feature interpretability.} 

Recently, more and more research works focus on compressing deep learning models as well as maintaining their performance. Given the fact that most of the trainable weights in Neural Network are in the tensor format and the operations in Neural Network can also be expressed as tensor operations, some research works tried to apply tensor decomposition techniques to project the weight tensors into a lower dimension space. For the weight tensor of the fully-connected layers (FC layers), a Tensor Train format (TT-format) was proposed to represent the weight tensor as the product of a series of lower-dimensional matrices (\cite{Novikov2015}). Tucker Tensor Layer (TTL) was proposed to employ Tucker Decomposition to decompose the weight tensor of FC layers into a core tensor and factor matrices (\cite{Calvi2019}). The main idea of these works is to approximate the high-dimensional weight tensor by using lower-dimensional matrices with fewer parameters. These methods can compress the neural networks well and save computational resources. However, they have three limitations: (i) they are designed to replace weight tensor in FC layers and they do not work for Conv layers. It is much more challenging to substitute decomposed components into Conv layers than FC layers. (ii) neither Tensor Train decomposition nor Tucker decomposition is unique. Intuitively and practically, uniqueness tests whether one decomposition can uncover the actual latent components rather than an arbitrarily rotated version. The non-uniqueness of TT-format and Tucker-format parameters limit the interpretability of the features. (iii) the decomposed components given by TT-format or Tucker Decomposition include three-way tensors which cannot be regarded as the convolutional kernel along a specific dimension of the input. Since the Conv layer of CNN  may generate a large number of unknown parameters, \cite{Leb15} {proposed to use CP-decomposition to decompose the weight tensor in the Conv layer to reduce the number of parameters. This method firstly trains a CNN model, and decomposes the weight tensor in the Conv layer, then replaces the original Conv layer with four stacked Conv layers and each of them is assigned a decomposed kernel as weights. The second step is to finetune the new CNN with four Conv layers on the training data. This method has three limitations: (i) it is an indirect training method, and it needs two-step training where CNN was trained multiple times. This may introduce extra computational cost; (ii) this method uses four new layers to replace the original Conv layer, which results in an increase in the depth of model; (iii) it lacks derivation of gradients with respect to (w.r.t). the decomposed weights and there  are still theoretical gaps in the analysis of adapting CP-decomposition into Conv layer.}

In this paper, we propose a novel layer to use CP-decomposition to approximately compress the Conv layer. {We name this new layer after CPAC-Conv layer. Different from Lebedev's method, our proposed CPAC-Conv layer does not add extra layers, so the depth of model remains the same}. We derive the expressions of both forward and backward propagation according to the first principles and matrix calculus. First, we derive the tensor expression of the Conv layer. Next, we substitute the CP-decomposition results of weight tensor into original expression and formulate the forward propagation expression of the CPAC-Conv layer. To {improve efficiency of parameters learning}, we derive the gradients w.r.t. each decomposed weight matrices to complete the backward propagation. To the best of our knowledge, our work is the first attempt to approximate and compress the Conv layer and derive the complete expressions of both the forward and backward propagations. {The contributions of this paper could be summarized as three aspects: (i) it creates a novel CPAC-Conv layer to approximate and compress the original Conv layer} {by decomposing the original convolutional kernel into a set of small kernels. Our proposed CPAC-Conv layer does not add extra layers and change the depth of the model.} {By using a smaller size of parameters and less model complexity, the proposed CPAC-Conv layer can achieve a comparable performance as the original Conv layers in deep learning models;  (ii) it provides complete theoretical derivations of the forward and backward propagations on the proposed CPAC-Conv layer, which lays a foundation to enable} {tuning each decomposed kernel independently}; {(iii) the proposed CPAC-Conv layer can achieve single-step CNN training with reduced parameters and model complexity, which can save computational time significantly. Besides, the reduced parameter size can enable better model generalization and provide insights on feature interpretation.}

The remainder of the paper will be organized as follows. Section \ref{background} introduces the theoretical backgrounds needed for method derivation, which include CP-decomposition and matrix calculus; Section \ref{forward-backward} creates the CPAC-Conv layer and contains the mathematical derivations of forward and backward propagations of CPAC-Conv layer; Section \ref{CPAC-CNN} introduces how to build up new CNN with CPAC-Conv layers (CPAC-CNN) and generates the training algorithm of CPAC-CNN; Section \ref{Case} shows the model performance {using the MNIST dataset} (\cite{lecun-mnisthandwrittendigit-2010}) {and the Magnetic Tile Defect dataset} (\cite{Huang2018}). {It also justifies the contributions of our work}; Section \ref{conclude} summarizes the conclusion of this paper.
\section{Theoretical Backgrounds}
\label{background}
Before diving into the detailed introduction of the CPAC-Conv layer, we first introduce the necessary theoretical backgrounds needed to derive and propose our method.

\subsection{CP-Decomposition}

The CANDECOMP/PARAFAC (CP) decomposition decomposes a tensor into a sum of rank-one tensors. For example, as shown in Figure 1, a three-way tensor $\mathcal{X} \in \mathbb{R}^{d\times d\times S}$ can be decomposed into the summation of $a_{i} \in \mathbb{R}^{d}, b_{i} \in \mathbb{R}^{d}, c_{i} \in \mathbb{R}^{S}, i = 1,...,R$. The expression of CP-decomposition is given {as the equation below}, where the operator $\circ$ represents outer product.
\begin{align}
    \mathcal{X} \approx \sum_{i=1}^{R} \mathbf{a}_{i} \circ \mathbf{b}_{i} \circ \mathbf{c}_{i}
\end{align}

\begin{figure}[H]
\centering
\resizebox*{12cm}{!}{\includegraphics{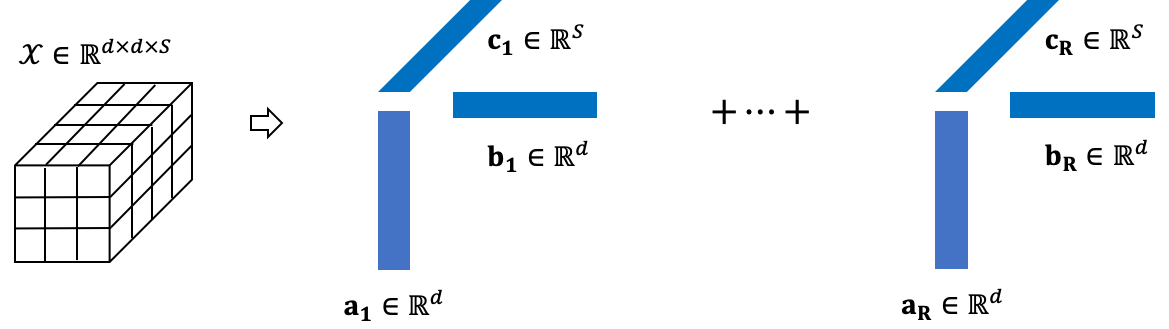}}
\caption{CP-decomposition of three-way tensor $\mathcal{X}$}
\label{CP_decomposition}
\end{figure}

Based on the CP-decomposition on three-way tensor, it can be further extended to a general $N_{th}$-order tensor, $\mathcal{X} \in \mathbb{R}^{I_{1} \times I_{2} \times ... \times I_{N}}$. The expression of general CP-decomposition is given as equation (\ref{3-1-1}), where the $\lambda_{r}$ represents the weights of each rank.
\begin{align}
    \mathcal{X} \approx \sum_{i=1}^{R} \lambda_{r} \mathbf{a}_{i}^{(1)} \circ \mathbf{a}_{i}^{(2)} \circ ... \circ \mathbf{a}_{i}^{(N)}
    \label{3-1-1}
\end{align}

CP-decomposition is one of the most popular tensor decompositions due to its intuitive interpretation and its uniqueness under very mild conditions. Practically, the uniqueness of extracted features indicates that the CP decomposition may uncover the latent factors and hidden patterns.  

\subsection{Matrix Calculus}
Matrix calculus is an important prerequisite in our proposed method. At first, the column stacking vectorization of a matrix $X \in \mathbb{R}^{m\times n}$ is given as equations (\ref{3-2-1}) (\cite{Crow89}).
\begin{align}
    \text{vec}(X) &= [X_{11},...,X_{m1}, X_{12},...,X_{m2}, ..., X_{1n},...,X_{mn}]^{T} \in \mathbb{R}^{mn \times 1}
    \label{3-2-1}
\end{align}
And then, there is an important relationship between the Kronecker product and the vectorization.
\begin{lemma}[\cite{Mag85}]
For any three matrices $A,X,B$, such that the matrix product $AXB$ is defined,
\end{lemma}
\begin{align}
    \text{vec}(AXB) = (B^{T} \otimes A)\text{vec}(X)
\end{align}
Furthermore, matrix derivative is essential in deriving back-propagation expressions of our proposed CPAC-Conv layer. Suppose we have matrix $F \in \mathbb{R}^{p \times q}$ and $X \in \mathbb{R}^{m \times n}$, we define the derivative of $F$ w.r.t. $X$ as equation (\ref{3-2-3}).
\begin{definition}[\cite{Mag85}]
Let $F$ be a differentiable $p\times q$ real matrix function of an $m\times n$ matrix of real variables $X$. The derivative of $F$ at $X$ is the $mn\times pq$ matrix.
\end{definition}
\begin{align}
    \frac{\partial F}{\partial X} = \frac{\partial \text{vec}(F)}{\partial\text{vec}(X)}
    \label{3-2-3}
\end{align}
We also have the extension of the first identification theorem, which describes the relationship between the vectorized matrix differential and derivative shown in equation (\ref{3-2-4}) (\cite{Crow89}).
\begin{align}
    \text{vec}(dF) = \frac{\partial F ^T}{\partial X}\text{vec}(dX)
    \label{3-2-4}
\end{align}

\subsection{{Convolutional Layer}}
\begin{figure}[H]
\centering
\resizebox*{11cm}{!}{\includegraphics{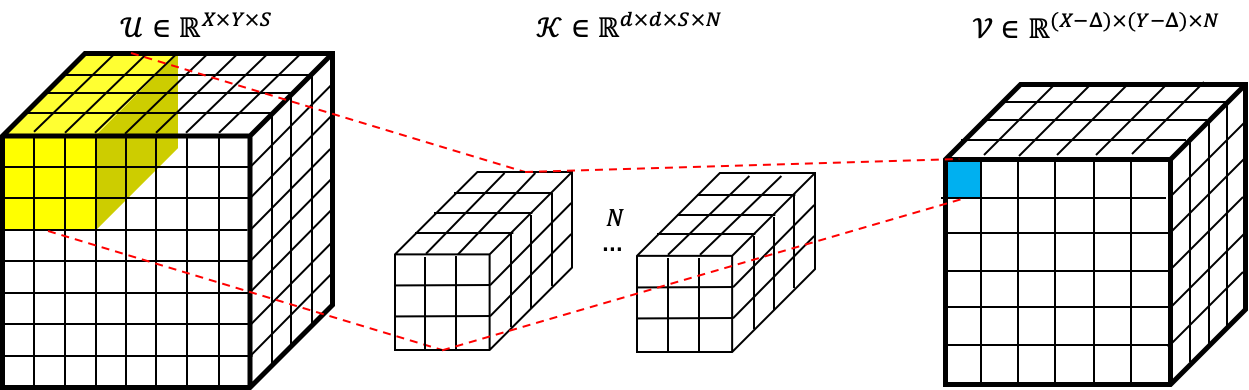}}
\caption{{Convolutional operation on input tensor $\mathcal{U}$}}
\label{Conv_layer}
\end{figure}

{The Conv layer is the fundamental component of CNN model, which basically conducts convolutional operation on input tensor $\mathcal{U} \in \mathbb{R}^{X\times Y \times S}$. The process of convolutional operation is shown in Figure }\ref{Conv_layer}, {and the corresponding expression is given in equation} (\ref{4-1-1}).
\begin{align}
    \mathcal{V}(x,y,n) = \sum_{i = x}^{x + d - 1}\sum_{j = y}^{y + d - 1}\sum_{s=1}^{S} \mathcal{K}(i-x,j-y,s,n)\mathcal{U}(i,j,s)
    \label{4-1-1}
\end{align}
{In equation} (\ref{4-1-1}), {$\mathcal{U} \in \mathbb{R}^{X\times Y \times S}$ denotes the input tensor, $X,Y$ represent width and height respectively and $S$ represents the number of channels; $\mathcal{K} \in \mathbb{R}^{d\times d\times S\times N}$ denotes the convolutional kernel, $d$ is the kernel size, and $N$ is the number of output channels. Given the stride is one and no padding applied, the output tensor should be $\mathcal{V} \in \mathbb{R}^{(X-\Delta) \times (Y-\Delta) \times N}$, in which $\Delta = d-1$, and $1$ is determined by stride.}

{In the convolutional operation, stride controls how the filter convolves over the input tensor. In equation }(\ref{4-1-1}), {we set stride as $1$, which means the filter will slide one pixel each time over the input tensor. Padding is another important technique in the convolutional operation. It is usually used to ensure that the input and output tensors share the same spatial dimension.} {Equation }(\ref{4-1-1}) {shows that the output of the Conv layer will shrink over spatial dimensions. The padding technique will be necessary for very deep CNN with many Conv layers, in which the final output may have severe shrinkage in spatial dimension after a series of convolution operations.}


\section{Adapt CP-decomposition to Approximately Compress Convolutional Layer}
\label{forward-backward}
In this section, we will create a novel CPAC-Conv layer by adapting CP-decomposition to approximately compress the Conv layer. In general, the basic idea of our proposed CPAC-Conv layer can be summarized into three aspects: (a) Since the high-rank tensor could be approximated by the summation of rank-one tensors, we {can} reduce the parameters in Conv layer by replacing the convolution kernel with a group of rank-one kernels given by CP-decomposition; (b) Since the accuracy of CP-decomposition is controlled by the hyper-parameter $R$ shown in equation (\ref{3-1-1}), {our proposed method can further improve its performance by tuning} $R$; (c) Convolution operations over those decomposed rank-one kernels can be regarded as a sequence of convolution operations along each axis of the input tensor. In this section, we will derive the expressions of forward and backward propagations of the CPAC-Conv layer, which {lays a theoretical foundation for developing and training new Deep Neural Networks with one or more CPAC-Conv layers}.
{The computational properties of the CPAC-Conv layer, including the time complexity and required memory, will also be discussed.}

\subsection{Forward Propagation of CPAC-Conv Layer}
\label{forward}
\begin{figure}
\centering
\resizebox*{12cm}{!}{\includegraphics{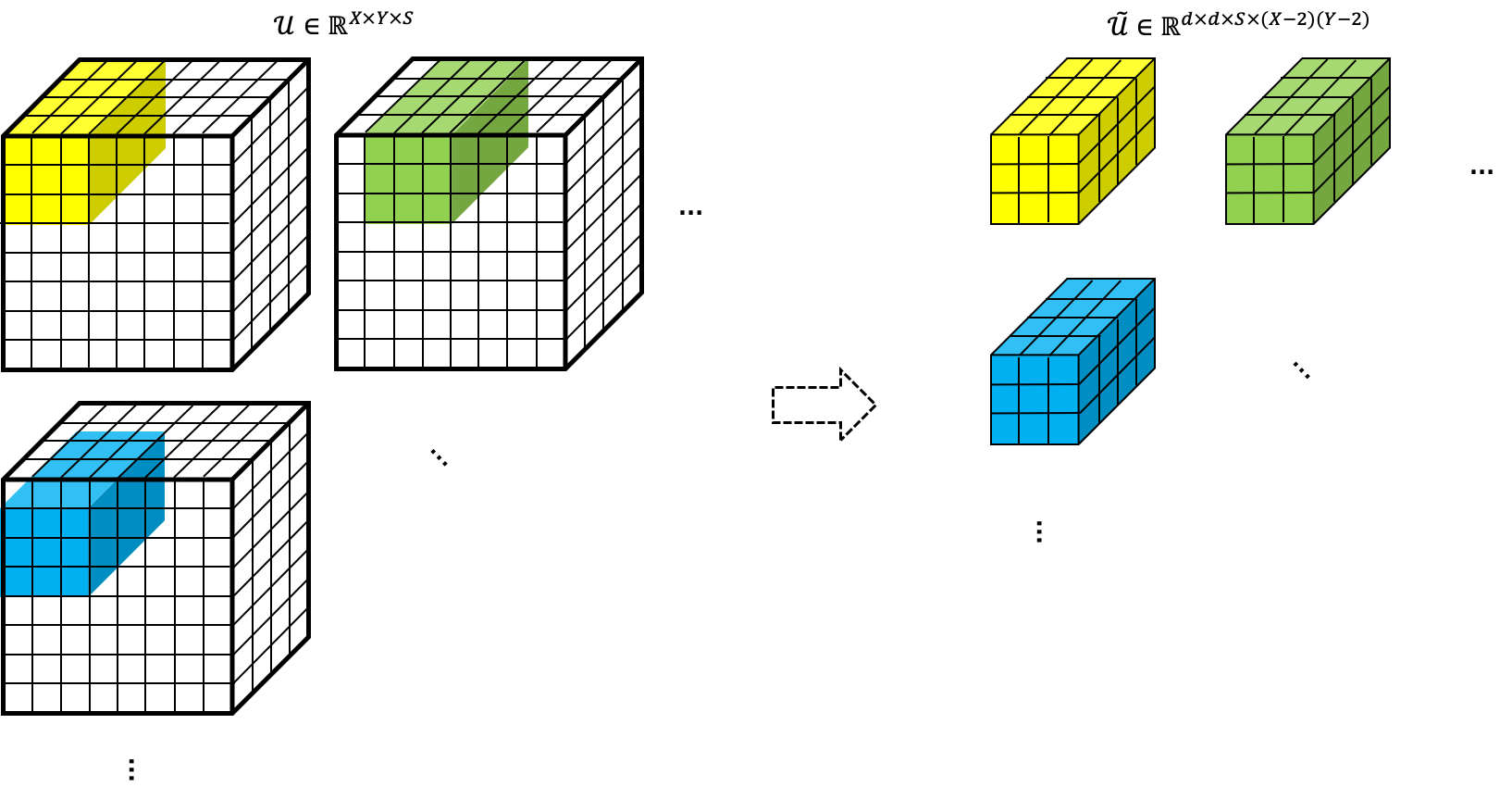}}
\caption{Visualization of Input Tensor Reshape}
\label{Input_Reshape}
\end{figure}

In this section, we will derive the forward propagation of CPAC-Conv layer based on the expression of original convolutional operation shown in equation (\ref{4-1-1}). To formulate the forward propagation of CPAC-Conv layer, we need at first apply CP-decomposition on the original convolutional kernel $\mathcal{K}$, which is a 4-way tensor with the shape of $d\times d\times S\times N$. The CP-decomposition of kernel $\mathcal{K}$ is given as equation (\ref{4-2-1}).
\begin{align}
    \mathcal{K} = \sum_{r=1}^R K^{X}_{r} \circ K^{Y}_{r} \circ K^{S}_{r} \circ K^{N}_{r}
    \label{4-2-1}
\end{align}
In equation (\ref{4-2-1}), the decomposed rank-one tensors, $K^{X}_{r}, K^{Y}_{r}, K^{S}_{r}, K^{N}_{r}, r = 1,...,R$, could be regarded as the group of small kernels along each axis of input tensor and the channels of output. After substituting the equation (\ref{4-2-1}) into equation (\ref{4-1-1}), the scalar expression of forward propagation can be obtained (\cite{Leb15}), as shown in equation (\ref{4-2-2}).
\begin{align}
    \mathcal{V}(x,y,n) = \sum_{r=1}^{R}K^{N}_{r}(n)\left(\sum_{i = x }^{x + d - 1}K^{X}_{r}(i-x)\left(\sum_{j = y}^{y + d - 1}K^{Y}_{r}(j-y)\left(\sum_{s=1}^{S}K^{S}_{r}(s)\mathcal{U}(i,j,s)\right)\right)\right)
    \label{4-2-2}
\end{align}
Next, we need to further reformulate the scalar expression (\ref{4-2-2}) into tensor expression. At first, we reshape the input $\mathcal{U} \in \mathbb{R}^{X\times Y \times S}$ into $\Tilde{\mathcal{U}} \in \mathbb{R}^{d\times d \times S\times (X-d+1)(Y-d+1)}$. The Figure \ref{Input_Reshape} shows an example of reshaping with $d=3$. Intuitively, as the original convolution operation is implemented by sliding the convolution kernel over the input tensor, the reshape process will duplicate and reorganize the input tensor to enable convolution operation finished by tensor production. The corresponding tensor expression of equation (\ref{4-2-2}) is shown in equation (\ref{4-2-3}).
\begin{align}
    \Tilde{\mathcal{V}} = \sum_{r=1}^R\left(\left(\left(\Tilde{\mathcal{U}} \times_{3} K_{r}^{S}\right) \times_{2} K_{r}^{Y} \right)\times_{1} K_{r}^{X}\right) \circ K_{r}^{N}
    \label{4-2-3}
\end{align}
The output $\Tilde{\mathcal{V}}$ is of shape $(X-d+1)(Y-d+1) \times N$, which can be transformed into the output of equation (\ref{4-1-1}) by one more step of reshaping. {From equation} (\ref{4-2-3}) {, the operation in our proposed CPAC-Conv layer can be regarded as conducting a series of original convolutional operations along each dimension of the input tensor and expanding to different channels of the output. In many applications of industrial engineering, each dimension of the tensor data has a specific physical meaning. Incorporating CP-decomposition into the convolutional kernel provides us with an opportunity to focus on features from specific dimensions that are more sensitive according to engineering prior knowledge on the input data.}
\subsection{Backward Propagation of CPAC-Conv Layer}
\label{backward}
{In} the previous section, we have derived the tensor expression of CPAC-Conv layer. Considering the simplest structure of a CNN model with CPAC-Conv layers ({named after CPAC-CNN for short}) as an example, the CPAC-CNN has one CPAC-Conv layer, and this layer is followed by one FC layer. So the output of the model can be expressed as equation (\ref{4-3-1}).
\begin{align}
    \hat{y} = \mathcal{W}\Tilde{\mathcal{V}} + b,
    \label{4-3-1}
\end{align}
in which, $\hat{y} \in \mathbb{R}^{C}$, $\mathcal{W} \in \mathbb{R}^{(X-d+1)(Y-d+1)\times N \times C}$, $b \in \mathbb{R}^{C}$, and $C$ is the number of class. The loss function is denoted as $\mathcal{L}(y,\hat{y})$, which represents the difference between the predicted labels $\hat{y}$ and the real labels $y$. To estimate the unknown parameters in Neural Network, stochastic gradient descent and its variants are applied by using back-propagation (\cite{Rum86}). To update the parameters in CPAC-Conv layer, {we need to derive the gradient of the loss function} $\mathcal{L}(y,\hat{y})$ w.r.t. the output $\hat{y}$, and then propagate the gradient backwards through each layer to derive the partial gradient w.r.t to each rank-one tensor in CPAC-Conv layer $(\frac{\partial \mathcal{L}}{\partial K_{r}^{N}}, \frac{\partial \mathcal{L}}{\partial K_{r}^{X}}, \frac{\partial \mathcal{L}}{\partial K_{r}^{Y}}, \frac{\partial \mathcal{L}}{\partial K_{r}^{S}})$. 
\subsubsection{Partial Derivative w.r.t. $K_{r}^{N}$}
According to the chain rule, the expression of $\frac{\partial \mathcal{L}}{\partial K_{r}^{N}}$ is shown in equation (\ref{4-3-2}), in which the first two components are the same as the original CNN models. {We need to derive} $\frac{\partial\Tilde{\mathcal{V}}}{\partial K_{r}^{N}}$.
\begin{align}
    \frac{\partial \mathcal{L}}{\partial K_{r}^{N}} = \frac{\partial \mathcal{L}}{\partial \hat{y}} \frac{\partial \hat{y}}{\partial\Tilde{\mathcal{V}}}\frac{\partial\Tilde{\mathcal{V}}}{\partial K_{r}^{N}}
    \label{4-3-2}
\end{align}
At first, we calculate the differential of equation (\ref{4-2-3}) at both sides w.r.t. $K_{r}^{N}$.
\begin{align}
    d\Tilde{\mathcal{V}} &= \left(\left(\left(\Tilde{\mathcal{U}} \times_{3} K_{r}^{S}\right) \times_{2} K_{r}^{Y} \right)\times_{1} K_{r}^{X}\right) \circ d K_{r}^{N}
    \label{4-3-3}
\end{align}
After vectorizing both sides, we have equation (\ref{4-3-4}).
\begin{align}
    \text{vec}(d\Tilde{\mathcal{V}}) &= \text{vec}\left(\left(\left(\left(\Tilde{\mathcal{U}} \times_{3} K_{r}^{S}\right) \times_{2} K_{r}^{Y} \right)\times_{1} K_{r}^{X}\right) \circ dK_{r}^{N}\right)
    \label{4-3-4}
\end{align}
We use $A_{1} \in \mathbb{R}^{(X-d+1)(Y-d+1) \times 1}$ to denote the constant part $\left(\left(\left(\Tilde{\mathcal{U}} \times_{3} K_{r}^{S}\right) \times_{2} K_{r}^{Y} \right)\times_{1} K_{r}^{X}\right)$, and $dK_{r}^{N} \in \mathbb{R}^{N \times 1}$. Substituting $A_{1}$ into equation (\ref{4-3-4}) yields

\begin{align}
    \text{vec}(d\Tilde{\mathcal{V}}) &= \text{vec}\left(A_{1} \circ dK_{r}^{N}\right)
    \notag\\
    &= \text{vec}\left(A_{1}  [dK_{r}^{N}]^{T}\right)
    \notag\\
    &= (I_{N} \otimes A_{1})\text{vec}\left([dK_{r}^{N}]^{T}\right)
    \notag\\
    &= (I_{N} \otimes A_{1})\text{vec}(dK_{r}^{N})
    \notag\\
    \frac{\text{vec}(d\Tilde{\mathcal{V}})}{\text{vec}(dK_{r}^{N})} &= (I_{N} \otimes A_{1})
    \notag\\
    \frac{\partial\Tilde{\mathcal{V}}}{\partial K_{r}^{N}} &= I_{N} \otimes A_{1}^{T}
    \label{4-3-5}
\end{align}
In equation (\ref{4-3-5}), because $dK_{r}^{N} \in \mathbb{R}^{N \times 1}$, we have $\text{vec}\left([dK_{r}^{N}]^{T}\right) = \text{vec}(dK_{r}^{N})$.
\subsubsection{Partial Derivative w.r.t. $K_{r}^{X}$}
The expression of $\frac{\partial \mathcal{L}}{\partial K_{r}^{X}}$ is shown in equation (\ref{4-3-6}), and we will derive $\frac{\partial\Tilde{\mathcal{V}}}{\partial K_{r}^{X}}$.
\begin{align}
    \frac{\partial \mathcal{L}}{\partial K_{r}^{X}} = \frac{\partial \mathcal{L}}{\partial \hat{y}} \frac{\partial \hat{y}}{\partial\Tilde{\mathcal{V}}}\frac{\partial\Tilde{\mathcal{V}}}{\partial K_{r}^{X}}
    \label{4-3-6}
\end{align}
At first, we calculate the differential of equation (\ref{4-2-3})  w.r.t. $K_{r}^{X}$. 
\begin{align}
    d\Tilde{\mathcal{V}} &= \left(\left(\left(\Tilde{\mathcal{U}} \times_{3} K_{r}^{S}\right) \times_{2} K_{r}^{Y} \right)\times_{1} d K_{r}^{X}\right) \circ K_{r}^{N}
    \label{4-3-7}
\end{align}
After vectorizing both sides, we have equation (\ref{4-3-8}).
\begin{align}
    \text{vec}(d\Tilde{\mathcal{V}}) &= \text{vec}\left(\left(\left(\left(\Tilde{\mathcal{U}} \times_{3} K_{r}^{S}\right) \times_{2} K_{r}^{Y} \right)\times_{1} d K_{r}^{X}\right) \circ K_{r}^{N}\right)
    \label{4-3-8}
\end{align}
We use $A_{2} \in \mathbb{R}^{d \times (X-d+1)(Y-d+1)}$ to denote the constant part $\left(\left(\Tilde{\mathcal{U}} \times_{3} K_{r}^{S}\right) \times_{2} K_{r}^{Y} \right)$, and $B_{1} \in \mathbb{R}^{N \times 1}$ to denote the constant part $K_{r}^{N}$. After substituting $A_{2}$ and $B_{1}$ into equation (\ref{4-3-8}), we can simplify it into equation (\ref{4-3-9}).
\begin{align}
    \text{vec}(d\Tilde{\mathcal{V}}) &= \text{vec}\left(A_{2}\times_{1} d K_{r}^{X} \circ B_{1}\right)
    \notag\\
    &= \text{vec}\left(A_{2}^{T} d K_{r}^{X} B_{1}^{T}\right)
    \notag\\
    &= (B_{1} \otimes A_{2}^{T})\text{vec}( d K_{r}^{X} )
    \notag\\
    \frac{\text{vec}(d\Tilde{\mathcal{V}})}{\text{vec}( d K_{r}^{X} )} &= (B_{1} \otimes A_{2}^{T})
    \notag\\
    \frac{\partial\Tilde{\mathcal{V}}}{\partial K_{r}^{X}} &= B_{1}^{T} \otimes A_{2}
    \label{4-3-9}
\end{align}
\subsubsection{Partial Derivative w.r.t. $K_{r}^{Y}$}
The expression of $\frac{\partial \mathcal{L}}{\partial K_{r}^{Y}}$ is shown in equation (\ref{4-3-10}), and we can derive $\frac{\partial\Tilde{\mathcal{V}}}{\partial K_{r}^{Y}}$.
\begin{align}
    \frac{\partial \mathcal{L}}{\partial K_{r}^{Y}} = \frac{\partial \mathcal{L}}{\partial \hat{y}} \frac{\partial \hat{y}}{\partial\Tilde{\mathcal{V}}}\frac{\partial\Tilde{\mathcal{V}}}{\partial K_{r}^{Y}}
    \label{4-3-10}
\end{align}
We calculate the differential of equation (\ref{4-2-3})  w.r.t. $K_{r}^{Y}$. 
\begin{align}
    d\Tilde{\mathcal{V}} &= \left(\left(\left(\Tilde{\mathcal{U}} \times_{3} K_{r}^{S}\right) \times_{2} d K_{r}^{Y} \right)\times_{1}  K_{r}^{X}\right) \circ K_{r}^{N}
    \label{4-3-11}
\end{align}
We use $\mathcal{A} \in \mathbb{R}^{d\times d\times (X-d+1)(Y-d+1)}$ to denote the constant part $\left(\Tilde{\mathcal{U}} \times_{3} K_{r}^{S}\right)$. After substituting it into equation (\ref{4-3-11}), we can get equation (\ref{4-3-12}).
\begin{align}
    d\Tilde{\mathcal{V}} &= \left(\left(\mathcal{A} \times_{2} d K_{r}^{Y} \right)\times_{1}  K_{r}^{X}\right) \circ K_{r}^{N}
    \label{4-3-12}
\end{align}
Then, we can use $\mathcal{A}_{(2)} \in \mathbb{R}^{d \times d(X-d+1)(Y-d+1)}$ to denote the mode$-2$ unfolding of $\mathcal{A}$. So we can rewrite the inner part $\left(\mathcal{A} \times_{2} d K_{r}^{Y} \right) \in \mathbb{R}^{d\times(X-d+1)(Y-d+1)}$ of equation (\ref{4-3-12}) into $P\left((\mathcal{A}_{(2)})^{T} dK_{r}^{Y}\right)\in \mathbb{R}^{(X-d+1)(Y-d+1)\times d}$. $P(.)$ represents a permutation operator, which means $(\mathcal{A}_{(2)})^{T} dK_{r}^{Y}$ and $\mathcal{A} \times_{2} d K_{r}^{Y}$ are matrices containing the same elements, but arranged differently. Equation (\ref{4-3-12}) can be transformed into equation (\ref{4-3-13}). 
\begin{align}
    d\Tilde{\mathcal{V}} &= \left[P\left((\mathcal{A}_{(2)})^{T} dK_{r}^{Y}\right)K_{r}^{X}\right] (K_{r}^{N})^{T}
    \label{4-3-13}
\end{align}
Further, we can use $B_{2} \in \mathbb{R}^{d \times N}$ to denote $K_{r}^{X}(K_{r}^{N})^{T}$. {By substituting} $B_{2}$ into equation (\ref{4-3-13}) {and vectorizing both sides}, we can get equation (\ref{4-3-14}).
\begin{align}
    \text{vec}(d\Tilde{\mathcal{V}}) &= \text{vec} \left(P\left((\mathcal{A}_{(2)})^{T} dK_{r}^{Y}\right)B_{2}\right)
    \notag\\
    &= B_{2}^{T} \otimes I_{(X-d+1)(Y-d+1)}\text{vec} \left(P\left((\mathcal{A}_{(2)})^{T} dK_{r}^{Y}\right)\right)
    \label{4-3-14}
\end{align}
Because $P(.)$ is a permutation operator, we can show that $\text{vec} \left(P\left((\mathcal{A}_{(2)})^{T} dK_{r}^{Y}\right)\right)$ is the same as $\text{vec} \left((\mathcal{A}_{(2)})^{T} dK_{r}^{Y}\right)$. 
\begin{align}
    \text{vec}(d\Tilde{\mathcal{V}}) &= B_{2}^{T} \otimes I_{(X-d+1)(Y-d+1)}\text{vec} \left(P\left((\mathcal{A}_{(2)})^{T} dK_{r}^{Y}\right)\right)
    \notag\\
    &= B_{2}^{T} \otimes I_{(X-d+1)(Y-d+1)}\text{vec} \left((\mathcal{A}_{(2)})^{T} dK_{r}^{Y}\right)
    \notag\\
    &= (B_{2}^{T} \otimes I_{(X-d+1)(Y-d+1)})(I_{1}\otimes\mathcal{A}_{(2)}^{T}) \text{vec} \left( dK_{r}^{Y}\right)
    \notag\\
    \frac{\text{vec}(d\Tilde{\mathcal{V}})}{\text{vec} \left( dK_{r}^{Y}\right)} &= (B_{2}^{T} \otimes I_{(X-d+1)(Y-d+1)})(I_{1}\otimes\mathcal{A}_{(2)}^{T})
    \notag\\
    \frac{\partial \Tilde{\mathcal{V}}}{\partial  K_{r}^{Y}} &= [(B_{2}^{T} \otimes I_{(X-d+1)(Y-d+1)})(I_{1}\otimes\mathcal{A}_{(2)}^{T})]^{T}
    \notag\\
    &= (I_{1}\otimes\mathcal{A}_{(2)})^{T}(B_{2}^{T} \otimes I_{(X-d+1)(Y-d+1)})^{T}
    \notag\\
    &= \mathcal{A}_{(2)}(B_{2} \otimes I_{(X-d+1)(Y-d+1)})
    \label{4-3-15}
\end{align}
\subsubsection{Partial Derivative w.r.t. $K_{r}^{S}$}
The expression of $\frac{\partial \mathcal{L}}{\partial K_{r}^{S}}$ is shown in equation (\ref{4-3-16}), and we will derive $\frac{\partial\Tilde{\mathcal{V}}}{\partial K_{r}^{S}}$.
\begin{align}
    \frac{\partial \mathcal{L}}{\partial K_{r}^{S}} = \frac{\partial \mathcal{L}}{\partial \hat{y}} \frac{\partial \hat{y}}{\partial\Tilde{\mathcal{V}}}\frac{\partial\Tilde{\mathcal{V}}}{\partial K_{r}^{S}}
    \label{4-3-16}
\end{align}
We firstly calculate the differential of equation (\ref{4-2-3}) w.r.t. $K_{r}^{S}$. 
\begin{align}
    d\Tilde{\mathcal{V}} &= \left(\left(\left(\Tilde{\mathcal{U}} \times_{3} d K_{r}^{S}\right) \times_{2}  K_{r}^{Y} \right)\times_{1}  K_{r}^{X}\right) \circ K_{r}^{N}
    \label{4-3-17}
\end{align}
In equation (\ref{4-3-17}), we have $\Tilde{\mathcal{U}} \in \mathbb{R}^{d\times d \times S \times (X-d+1)(Y-d+1)}$. We can use $\Tilde{\mathcal{U}}_{(3)} \in \mathbb{R}^{S \times d^{2}(X-d+1)(Y-d+1)}$ to represent the mode-3 unfolding of tensor $\Tilde{\mathcal{U}}$. Similarly, we use $P(.)$ to represent permutation operator and $(\Tilde{\mathcal{U}}_{(3)}^{T} d K_{r}^{S}) \in \mathbb{R}^{d^{2}(X-d+1)(Y-d+1)}$, $P(\Tilde{\mathcal{U}}_{(3)}^{T} d K_{r}^{S}) \in \mathbb{R}^{d(X-d+1)(Y-d+1) \times d}$, $P\left(P(\Tilde{\mathcal{U}}_{(3)}^{T} d K_{r}^{S}) K_{r}^{Y}\right) \in \mathbb{R}^{(X-d+1)(Y-d+1) \times d}$. So $\left(\left(\Tilde{\mathcal{U}} \times_{3} d K_{r}^{S}\right) \times_{2}  K_{r}^{Y}\right) \in \mathbb{R}^{d\times (X-d+1)(Y-d+1)}$ in the equation (\ref{4-3-17}) can be transformed into equation $\left[P\left(P(\Tilde{\mathcal{U}}_{(3)}^{T} d K_{r}^{S}) K_{r}^{Y}\right)\right]^{T} \in \mathbb{R}^{d \times (X-d+1)(Y-d+1)}$.
\begin{align}
    d\Tilde{\mathcal{V}} &= \left(\left[P\left(P(\Tilde{\mathcal{U}}_{(3)}^{T} d K_{r}^{S}) K_{r}^{Y}\right)\right]^{T}\times_{1}  K_{r}^{X}\right) \circ K_{r}^{N}
    \notag\\
    & = \left(P\left(P(\Tilde{\mathcal{U}}_{(3)}^{T} d K_{r}^{S}) K_{r}^{Y}\right)K_{r}^{X}\right)  (K_{r}^{N})^{T}
    \label{4-3-18}
\end{align}
After vectorizing both sides in equation (\ref{4-3-18}).
\begin{align}
    \text{vec} (d\Tilde{\mathcal{V}})
    & = \text{vec}\left(P\left(P(\Tilde{\mathcal{U}}_{(3)}^{T} d K_{r}^{S}) K_{r}^{Y}\right)K_{r}^{X}  (K_{r}^{N})^{T}\right)
    \notag\\
    & = B_{2}^{T} \otimes I_{(X-d+1)(Y-d+1)}\text{vec}\left(P\left(P(\Tilde{\mathcal{U}}_{(3)}^{T} d K_{r}^{S}) K_{r}^{Y}\right) \right)
    \notag\\
    & = B_{2}^{T} \otimes I_{(X-d+1)(Y-d+1)}\text{vec}\left(P(\Tilde{\mathcal{U}}_{(3)}^{T} d K_{r}^{S}) K_{r}^{Y}\right)
    \notag\\
    & = \left(B_{2}^{T} \otimes I_{(X-d+1)(Y-d+1)}\right)\left((K_{r}^{Y})^{T} \otimes I_{d(X-d+1)(Y-d+1)}\right)\text{vec}\left(P(\Tilde{\mathcal{U}}_{(3)}^{T} d K_{r}^{S})\right)
    \notag\\
    & = \left(B_{2}^{T} \otimes I_{(X-d+1)(Y-d+1)}\right)\left((K_{r}^{Y})^{T} \otimes I_{d(X-d+1)(Y-d+1)}\right)\text{vec}\left(\Tilde{\mathcal{U}}_{(3)}^{T} d K_{r}^{S}\right)
    \notag\\
    & = \left(B_{2}^{T} \otimes I_{(X-d+1)(Y-d+1)}\right)\left((K_{r}^{Y})^{T} \otimes I_{d(X-d+1)(Y-d+1)}\right)\left(I_{1} \otimes \Tilde{\mathcal{U}}_{(3)}^{T}\right)\text{vec}\left(d K_{r}^{S}\right)
    \notag\\
    \frac{\text{vec} (d\Tilde{\mathcal{V}})}{\text{vec}\left(d K_{r}^{S}\right)} &= \left(B_{2}^{T} \otimes I_{(X-d+1)(Y-d+1)}\right)\left((K_{r}^{Y})^{T} \otimes I_{d(X-d+1)(Y-d+1)}\right)\left(I_{1} \otimes \Tilde{\mathcal{U}}_{(3)}^{T}\right)
    \notag\\
    \frac{\partial \Tilde{\mathcal{V}}}{\partial K_{r}^{S}} &= \left[\left(B_{2}^{T} \otimes I_{(X-d+1)(Y-d+1)}\right)\left((K_{r}^{Y})^{T} \otimes I_{d(X-d+1)(Y-d+1)}\right)\Tilde{\mathcal{U}}_{(3)}^{T}\right]^{T}
    \notag\\
    \frac{\partial \Tilde{\mathcal{V}}}{\partial K_{r}^{S}} &= \Tilde{\mathcal{U}}_{(3)}\left(K_{r}^{Y} \otimes I_{d(X-d+1)(Y-d+1)}\right) \left(B_{2} \otimes I_{(X-d+1)(Y-d+1)}\right)
    \label{4-3-19}
\end{align}
Combining equations (\ref{4-3-5},\ref{4-3-9},\ref{4-3-15},\ref{4-3-19}), we generated the detailed expressions of $\{\frac{\partial\Tilde{\mathcal{V}}}{\partial K_{r}^{N}},\frac{\partial\Tilde{\mathcal{V}}}{\partial K_{r}^{X}}, \frac{\partial\Tilde{\mathcal{V}}}{\partial K_{r}^{Y}}, \frac{\partial\Tilde{\mathcal{V}}}{\partial K_{r}^{S}}\}$, which will be used to update decomposed kernels of CPAC-Conv layer in the backward propagation. 

{Up to now, we have derived the forward and backward propagations of CPAC-Conv layer. In the following Section} \ref{CPAC-CNN},{ we will discuss how to use the proposed CPAC-Conv layer to compress the parameters and reduce model complexity in Convolutional Neural Network.}

\section{Convolutional Neural Network with CPAC-Conv Layer (CPAC-CNN)}
\label{CPAC-CNN}
{In this section, we will give a general setup of CPAC-CNN and analyze computational properties of CPAC-Conv layer. It is worth noting that the proposed CPAC-Conv layer is not only for Convolutional Neural Network (CNN), it is also applicable to other deep learning models} {with the Conv layer as the building block, such as the Conv-LSTM model used in the spatial-temporal forecast, the Auto-encoder used in unsupervised learning, etc.}.
\subsection{General Setup of CPAC-CNN}
Suppose the CPAC-CNN consists of $L$ layers, among them, the first $(L-1)$ layers are CPAC-Conv layers, and the $L_{th}$ layer (output layer) is a fully-connected (FC) layer. Similar to section \ref{forward}, we use $\Tilde{\mathcal{U}}^{(l)}$ and $\Tilde{\mathcal{V}}^{(l)}$ to represent the input and output of the $l_{th}$ layer, in which $\Tilde{\mathcal{U}}^{(l)} \in \mathbb{R}^{d\times d\times N_{l-1}\times (X-l(d-1))(Y-l(d-1))},\Tilde{\mathcal{V}}^{(l)} \in \mathbb{R}^{(X-l(d-1))(Y-l(d-1))\times N_{l}}, l=1,...,L-1$. The detailed model setup and notations can be found in Table \ref{model-setup}.
\begin{table}
\newcommand{\tabincell}[2]{\begin{tabular}{@{}#1@{}}#2\end{tabular}}
\tbl{CPAC-CNN Setup and Notations}
{\begin{tabular}{ccccc} \toprule
 Layer Index & Layer Name & $\tabincell{c}{Weights \\ (Rank)}$ & $\tabincell{c}{Input \\ (Shape)}$ & Output \\ \midrule
 $1$  & CPAC-Conv &\tabincell{c}{$K^{(1)X}_{r}, K^{(1)Y}_{r}, K^{(1)S}_{r}, K^{(1)N_{1}}_{r}$\\ $(r = 1,...,R)$}  & $\tabincell{c}{$\Tilde{\mathcal{U}}^{(1)}$ \\ $( \mathbb{R}^{d\times d\times S\times (X-(d-1))(Y-(d-1))})$}$ & $\tabincell{c}{$\Tilde{\mathcal{V}}^{(1)} $ \\ ($ \mathbb{R}^{(X-(d-1))(Y-(d-1))\times N_{1}}$)}$  \\
 $2,...,L-1$ & CPAC-Conv &\tabincell{c}{$K^{(l)X}_{r}, K^{(l)Y}_{r}, K^{(l)N_{l-1}}_{r}, K^{(l)N_{l}}_{r}$\\ $(r = 1,...,R)$}  & $\tabincell{c}{$\Tilde{\mathcal{U}}^{(l)} $ \\ ($ \mathbb{R}^{d\times d\times N_{l-1}\times (X-l(d-1))(Y-l(d-1))}$)}$ & $\tabincell{c}{$\Tilde{\mathcal{V}}^{(l)} $ \\ ($ \mathbb{R}^{(X-l(d-1))(Y-l(d-1))\times N_{l}}$)}$  \\
 $L$ & FC & $\mathcal{W}^{(L)},b$ & $\tabincell{c}{$\Tilde{\mathcal{U}}^{(L)} $ \\ ($ \mathbb{R}^{(X-(L-1)(d-1))(Y-(L-1)(d-1))\times N_{L-1}})$}$ & $\tabincell{c}{$\hat{y}$ \\ ($ \mathbb{R}^{C}$)}$  \\
 \bottomrule
\end{tabular}}
\label{model-setup}
\end{table}

Given the general CPAC-CNN setup with $(L-1)$ CPAC-Conv layers and one FC layer, we further summarize the back-propagation expressions of CPAC-CNN according to the derivations in section \ref{backward}. For an arbitrary $l_{th} ( l=1,...,L-1)$ CPAC-Conv layer, the gradients of the loss function w.r.t. the weight matrices are computed as equations (\ref{5-1}).
\begin{align}
    \frac{\partial \mathcal{L}}{\partial K_{r}^{(l)X}} &= \frac{\partial \mathcal{L}}{\partial\Tilde{\mathcal{V}}^{(l)}}\frac{\partial\Tilde{\mathcal{V}}^{(l)}}{\partial K_{r}^{(l)X}}, r=1,...,R
    \notag\\
    \frac{\partial \mathcal{L}}{\partial K_{r}^{(l)Y}} &= \frac{\partial \mathcal{L}}{\partial\Tilde{\mathcal{V}}^{(l)}}\frac{\partial\Tilde{\mathcal{V}}^{(l)}}{\partial K_{r}^{(l)Y}}, r=1,...,R
    \notag\\
    \frac{\partial \mathcal{L}}{\partial K_{r}^{(l)N_{l-1}}} &= \frac{\partial \mathcal{L}}{\partial\Tilde{\mathcal{V}}^{(l)}}\frac{\partial\Tilde{\mathcal{V}}^{(l)}}{\partial K_{r}^{(l)N_{l-1}}}, r=1,...,R
    \notag\\
    \frac{\partial \mathcal{L}}{\partial K_{r}^{(l)N_{l}}} &= \frac{\partial \mathcal{L}}{\partial\Tilde{\mathcal{V}}^{(l)}}\frac{\partial\Tilde{\mathcal{V}}^{(l)}}{\partial K_{r}^{(l)N_{l}}}, r=1,...,R
    \label{5-1}
\end{align}
In equations (\ref{5-1}), the $\frac{\partial \mathcal{L}}{\partial\Tilde{\mathcal{V}}^{(l)}}$ is the gradient of the loss function w.r.t. the output of $l_{th}$, which is back propagated from the $(l+1)_{th}$ layer, and the detailed expressions of $\{\frac{\partial\Tilde{\mathcal{V}}^{(l)}}{\partial K_{r}^{(l)X}}, \frac{\partial\Tilde{\mathcal{V}}^{(l)}}{\partial K_{r}^{(l)Y}}, \frac{\partial\Tilde{\mathcal{V}}^{(l)}}{\partial K_{r}^{(l)N_{l-1}}}, \frac{\partial\Tilde{\mathcal{V}}^{(l)}}{\partial K_{r}^{(l)N_{l}}}\}$ are in equations (\ref{4-3-9},\ref{4-3-15},\ref{4-3-19},\ref{4-3-5}). Thus, we can summarize the training process of CPAC-CNN as Algorithm \ref{algorithm}. {Without loss of generality, we suppose the rank $R$ are selected to be same for all layers} $1,...,L-1$ {for simplicity. In addition, the choice of the activation function is not explicitly specified in Algorithm 1. The activation function is directly applied to the output of each layer and can be adaptively selected according to the requirements of specific tasks. The choice of the activation function does not influence the derivation of forward and backward propagations for our proposed CPAC-Conv layer. } {In this paper, we focus on compressing one Conv layer by using the proposed CPAC-Conv layer. Since we use a regular CNN model to compare the performances between our proposed CPAC-Conv layer and the original Conv layer, the padding technique is not used. }

\begin{algorithm}[ht]
  \caption{Forward and Backward-Propagation for CPAC-CNN}
  \begin{algorithmic}[1]
    \Inputs{$\mathcal{U} \in \mathbb{R}^{X\times Y\times S}, y$}
    \Initialize{$K^{(l)X}_{r}, K^{(l)Y}_{r}, K^{(l)N_{l-1}}_{r}, K^{(l)N_{l}}_{r} (N_{0}=S;r=1,...,R;l=1,...,L-1)$\\
    $\Tilde{\mathcal{W}}^{(L)}, b$\\
    $\Tilde{\mathcal{U}}^{(1)} \gets$  reshape the input $\mathcal{U}$}
    \State{\textbf{Forward Propagation:}}
    \For{$l = 1$ to $L-1$}
      \State $\Tilde{\mathcal{V}}^{(l)} \gets \sum_{r=1}^R\left(\left(\left(\Tilde{\mathcal{U}}^{(l)}\times_{3} K_{r}^{(l)N_{l-1}}\right) \times_{2} K_{r}^{(l)Y} \right)\times_{1} K_{r}^{(l)X}\right) \circ K_{r}^{(l)N_{l}}$
      \If{$l\leq L-2$}
      \State $\Tilde{\mathcal{U}}^{(l+1)} \gets \text{reshape the } \Tilde{\mathcal{V}}^{(l)}$
      \EndIf
    \EndFor
    \State Output of Fully Connected layer $\hat{y} = \mathcal{W}^{(L)}\Tilde{\mathcal{V}}^{(L-1)} + b$
    \State Calculate loss function $\mathcal{L}(y,\hat{y})$
    \State{\textbf{Backward Propagation:}}
    \State Calculate $\{\frac{\partial \mathcal{L}}{\partial \hat{y}}, \frac{\partial \hat{y}}{\partial\Tilde{\mathcal{V}}^{(L-1)}},\frac{\partial \hat{y}}{\partial\Tilde{\mathcal{W}}^{(L)}}\}$\Comment{the same as original CNN}
    \State Update $\Tilde{\mathcal{W}}^{(L)}$
    \For{$l=L-1$ to $1$}
    \For{$r=1$ to $R$}
    \State Calculate $\{\frac{\partial\Tilde{\mathcal{V}}^{(l)}}{\partial K_{r}^{(l)X}}, \frac{\partial\Tilde{\mathcal{V}}^{(l)}}{\partial K_{r}^{(l)Y}}, \frac{\partial\Tilde{\mathcal{V}}^{(l)}}{\partial K_{r}^{(l)N_{l-1}}}, \frac{\partial\Tilde{\mathcal{V}}^{(l)}}{\partial K_{r}^{(l)N_{l}}}\}$\Comment{equations (\ref{4-3-9},\ref{4-3-15},\ref{4-3-19},\ref{4-3-5})}
    \State Update $K^{(l)X}_{r}, K^{(l)Y}_{r}, K^{(l)N_{l-1}}_{r}, K^{(l)N_{l}}_{r}$ \Comment{according to selected optimizer}
    \EndFor
    \EndFor
  \end{algorithmic}
  \label{algorithm}
\end{algorithm}
\subsection{Properties of CPAC-Conv Layer}
\label{Properties}
{In this section, we will further compare the properties of our proposed CPAC-Conv layer with the original Conv layer. The property analysis includes the number of parameters, time complexity, and required memory.}
\subsubsection{Analysis of the Number of Parameters}
\label{analysis of parameters}
Compared with original CNN, one important property of our proposed CPAC-CNN is that it could reduce the model parameters by setting different values of rank $R$ in CP-decomposition. For example, given a Conv layer with kernel $\mathcal{K} \in \mathbb{R}^{d\times d\times S\times N}$, the number of parameters in this layer is $M_{1}=d\times d\times S\times N$. In our proposed CPAC-Conv layer, we use a series of small kernels $K^{X}_{r}, K^{Y}_{r}, K^{S}_{r}, K^{N}_{r} (r=1,...,R)$ to approximate the original kernel. The number of parameters in these kernels is $M_{2}=R\times(d+d+S+N)$. To numerically analyze the compression effect of CPAC-Conv layer, we use compression ratio (CR) to denote the ratio of parameters in CPAC-Conv layers and Conv layers, which is expressed as $\text{CR} = \frac{M_{2}}{M_{1}}$. It is easy to see that $M_{2}$ is determined by the rank $R$ of CP-decomposition, and {usually we select an} $R$ which makes $M_{2} < M_{1}$. By controlling the value of $R$, our proposed CPAC-Conv layer could use fewer parameters to approximate a Conv layer and reduce the model complexity. The proposed CPAC-Conv layer can also be extended and integrated with other deep learning models. 

\subsubsection{{Analysis of Floating Point Operations}}
\label{time_complexity}

{The number of floating-point operations (FLOPs) is commonly used to evaluate the time complexity of a proposed neural network. The idea of the FLOPs analysis is to count the basic operations, such as addition, multiplication, etc. in the neural networks } (\cite{DBLP:journals/corr/MolchanovTKAK16}). {Suppose the original Conv layer has  input $\mathcal{U} \in \mathbb{R}^{X\times Y \times S}$,  convolutional kernel $\mathcal{K} \in \mathbb{R}^{d\times d\times S\times N}$, and  output $\mathcal{V} \in \mathbb{R}^{(X-d+1) \times (Y-d+1) \times N}$, then the number of FLOPs is shown in equation} (\ref{4-4-1}) .

\begin{align}
    \text{FLOPs}_{\text{conv}} = 2NSd^{2}(X-d+1)(Y-d+1)
    \label{4-4-1}
\end{align}

{Note that for simplicity, the padding and bias are not considered in this case. For our proposed CPAC-Conv layer, according to equation} (\ref{4-2-3}), {it can be regarded as a series of four convolutional operations. The number of FLOPs is given in equation} (\ref{4-4-2}). 

\begin{align}
    \text{FLOPs}_{\text{CPAC-Conv}} = 2R(N+2d+S)(X-d+1)(Y-d+1)
    \label{4-4-2}
\end{align}

{By comparing the results of equations }(\ref{4-4-1}) and (\ref{4-4-2}), {we can conclude that the number of FLOPs of the CPAC-Conv layer can be much smaller than the original Conv layer by selecting the suitable value of rank $R$.} 
\subsubsection{{Analysis of Required Memory}}

{The required memory of a model can be divided into two parts, one of which is the memory taken by model parameters, the other is the memory taken by intermediate variables. The intermediate variables include the output of each layer in forward propagation and the gradients of model parameters in backward propagation. Compared with the original Conv layer, our proposed CPAC-Conv layer could use fewer parameters which could not only reduce the memory taken by parameters but also reduce the memory taken by backward gradients.} 

{We can take the memory taken by parameters as an example and compare the Conv layer ($\mathcal{K} \in \mathbb{R}^{d\times d\times S\times N}$) with the corresponding CPAC-Conv layer ($K^{X}_{r}, K^{Y}_{r}, K^{S}_{r}, K^{N}_{r} (r=1,...,R)$). If we store the parameters as the single-precision float format, each parameter will take 32 bits in the computer memory which is 4 Bytes (B). According to the analysis in section }\ref{analysis of parameters}{ ,the number of parameters in the Conv layer is $M_{1} = d\times d \times S\times N$, which will take $4d^{2}SN\times 10^{-6}$ Megabytes (MB) memory, and the number of parameters in the corresponding CPAC-Conv layer is $M_{2}=R\times (d+d+S+N)$, which will take $4R(2d+S+N)\times 10^{-6}$ MB memory. To this extent, the parameters in the CPAC-Conv layer will take less memory compared with the corresponding Conv layer by selecting an appropriate value of $R$. Similarly, in the backpopagation, the gradients are calculated w.r.t. each parameters, which will also take memory. The reduction of parameters also reduces the number of required gradients so that the memory taken by gradients will reduce accordingly.}

In summary, the properties comparison of the proposed CPAC-Conv layer and the original Conv layer are concluded in Table \ref{property}.

\begin{table}
\newcommand{\tabincell}[2]{\begin{tabular}{@{}#1@{}}#2\end{tabular}}
\tbl{Properties Comparisons}
{\begin{tabular}{ccc} \toprule
& Conv Layer & CPAC-Conv Layer\\ \midrule
 Number of Parameters  & $d\times d\times S\times N$ & $R \times (d + d + S + N$) \\
 FLOPs & $2NSd^{2}(X-d+1)(Y-d+1)$ & $2R(N+2d+S)(X-d+1)(Y-d+1)$ \\
 Required Memory (MB) & $4d^{2}SN\times 10^{-6}$ & $4R(2d + S + N)\times 10^{-6}$\\
 \bottomrule
\end{tabular}}
\label{property}
\end{table}

\section{Case Study}
\label{Case}
In this section, we {evaluate our proposed CPAC-CNN model based on two datasets}: one  is classification on the MNIST dataset (\cite{lecun-mnisthandwrittendigit-2010}), and the other is defect diagnosis on the Magnetic Tile Defects dataset (\cite{Huang2018}). By comparing with the original CNN model, we would like to justify the strengths of our proposed method from two aspects: (1) our proposed CPAC-Conv layer can receive a comparable performance compared with original Conv layer by using fewer parameters; (2) the CPAC-Conv layer increases the interpretability of the output feature maps. The experiments are implemented by PyTorch (\cite{NIPS2019_9015}) on a single NVIDIA GeForce GTX 1080Ti GPU. {Rectified Linear Unit (ReLU), which is widely used in image classification, is selected as the activation function in our experiment}. The code is available on \url{https://github.com/wyn430/CPAC-CNN}.

\subsection{Case 1: MNIST Dataset}
\label{case1}

The MNIST dataset is originally collected for handwritten digit recognition, and we use it to test the performance of our model on classification. This dataset consists of $60000$ $28\times 28$ grayscale images for training, and $10000$ for testing. We test the CPAC-CNN with one CPAC-Conv layer and two CPAC-Conv layers, and then we use CNN with one Conv layer and two Conv layers as benchmarks, respectively. The hyper-parameter in our proposed CPAC-CNN is the rank $R$ of CP-Decomposition, which can change the the number of trainable parameters in the model. In Table \ref{MNIST_Results}, we summarize the experiment results and model information, which include model structure, rank $R$ of CP-Decomposition, kernel size of convolution operation, number of parameters in Conv/CPAC-Conv layers, compression ratio (CR), and classification accuracy. From Table \ref{MNIST_Results}, we can see that (1) As the increase of Rank ($R$), the CPAC-CNN tends to have more parameters, and receive a better classification accuracy; (2) {By introducing our proposed CPAC layer, we can find a $R$ to achieve comparable classification performance with fewer parameters and reduced model complexity.}
\begin{table}[htp]
\newcommand{\tabincell}[2]{\begin{tabular}{@{}#1@{}}#2\end{tabular}}
\tbl{Experiment Results on MNIST}
{\begin{tabular}{ccccccc} \toprule
 Model & Structure & Rank (R) & Kernel Size & \# of Parameters& CR\textsuperscript{*} & Accuracy \\
 \midrule
 CNN &  \tabincell{c}{1$\times$Conv Layer \\ 1$\times$FC Layer}& - & $(8,3,3,1)$ & 72 & 1 & 0.9794\\
  \multirow{6}{*}{CPAC-CNN}&  \multirow{6}{*}{\tabincell{c}{1$\times$CPAC-Conv Layer \\ 1$\times$FC Layer}}& 1 & \multirow{6}{*}{$\tabincell{c}{$(8,R)$ \\ $(3,R)$ \\ $(3,R)$ \\ $(1,R)$}$} & \multirow{6}{*}{$15R$} & 0.2083 &0.9518 \\
  &  & 2 &  &  & 0.4167 & 0.9669\\
  &  & 3 &  &  & 0.6250 & 0.9677\\
  &  & 4 &  &  & 0.8333 & 0.9773\\
  &  & 5 &  &  & 1.0417 & 0.9782\\
  &  & 6 &  &  & 1.2500 & 0.9782\\
 \midrule
 CNN & \tabincell{c}{2$\times$Conv Layer \\ 1$\times$FC Layer} & - & $(8,3,3,1),(8,3,3,8)$ & 648 & 1 & 0.9844\\
  \multirow{12}{*}{CPAC-CNN}& \multirow{12}{*}{\tabincell{c}{2$\times$CPAC-Conv Layer \\ 1$\times$FC Layer}} & 1 & \multirow{12}{*}{$\tabincell{c}{$(8,R),(8,R)$ \\ $(3,R),(3,R)$ \\ $(3,R),(3,R)$ \\ $(1,R),(8,R)$}$} & \multirow{12}{*}{$37R$} & 0.0571  & 0.9278 \\
  &  & 2 &  &  & 0.1142  & 0.9679\\
  &  & 3 &  &  & 0.1713  & 0.9774\\
  &  & 4 &  &  & 0.2284  & 0.9788\\
  &  & 5 &  &  & 0.2855  & 0.9805\\
  &  & 6 &  &  & 0.3426  & 0.9767\\
  &  & 7 &  &  & 0.3997  & 0.9801\\
  &  & 8 &  &  & 0.4568  & 0.9800\\
  &  & 9 &  &  & 0.5139  & 0.9830\\
  &  & 10 &  &  & 0.5710  & 0.9833\\
  &  & 11 &  &  & 0.6281  & 0.9815\\
  &  & 12 &  &  & 0.6852 & 0.9842\\
 \bottomrule
\end{tabular}}
\tabnote{\textsuperscript{*}CR: Compression Ratio}
\label{MNIST_Results}
\end{table}

\begin{figure}[htp]
\centering
\resizebox*{14cm}{!}{\includegraphics{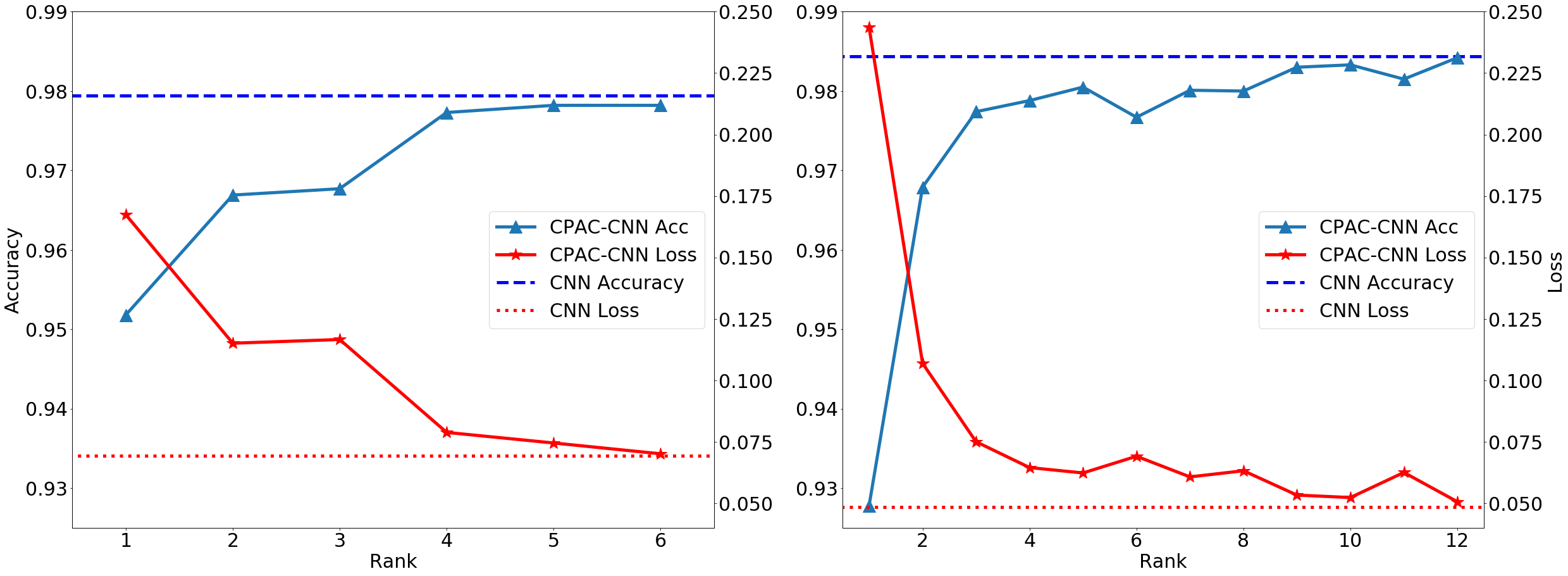}}
\caption{Performance Comparison Between CNN and CPAC-CNN on MNIST, \textbf{Left:} one Conv/CPAC-Conv Layer, \textbf{Right:} two Conv/CPAC-Conv Layers}
\label{MNIST_Results_Viz}
\end{figure}

For better visualization, we show the change of classification accuracy and loss along with various values of rank in Figure \ref{MNIST_Results_Viz}. The left plot in Figure \ref{MNIST_Results_Viz} is the result of single-layer models and the right plot is the result of double-layer models. The dashed line in Figure \ref{MNIST_Results_Viz} represents the classification accuracy of CNN, and the dotted line represents the loss function value of CNN. The line with triangle markers represents the change of classification accuracy of CPAC-CNN with various ranks, and the line with star markers represents the change of loss function value of CPAC-CNN with various ranks. The horizontal axis represents different ranks, the left vertical axis represents classification accuracy, and the right vertical axis represents the value of loss function. From Figure \ref{MNIST_Results_Viz}, we can find that with a larger rank ($R$), the CPAC-CNN receives better performance and gradually approaches the corresponding CNN. Considering the results shown in Table \ref{MNIST_Results} and Figure \ref{MNIST_Results_Viz}, we can conclude that (1) compared with original CNN, our proposed CPAC-CNN could receive a comparable classification accuracy with fewer parameters and reduced model complexity; (2) the classification accuracy and loss value tend to improve with the increasing of rank $R$; since the rank $R$ determines the number of parameters in CPAC-CNN, we can tune the rank $R$ to meet the accuracy requirement and satisfy the computation resource limitation; (3) the compression effect will be more significant as the model contains more CPAC-Conv layers.

Apart from comparing the number of parameters and performances with the original Conv layer, the computational time of our proposed CPAC-Conv layer and the original Conv layer are further compared. The theoretical analysis of the time complexity of these two methods is discussed in section \ref{time_complexity}. Considering the overall training time is influenced by the number of training samples, batch size, and the number of training epochs, without loss of generality, we compare the ``unit processing time", which is the time consumption to conduct a pair of forward and backward propagations on a single image. For the single-layer models, the ``unit processing time" of the original CNN and the CPAC-CNN are $0.002$ and $0.004$ seconds, respectively. For the double-layer models, the ``unit processing time" of the original CNN and the CPAC-CNN are $0.004$ and $0.009$ seconds, respectively. The rank of the aforementioned CPAC-CNN models is $1$. For the single-layer and double-layer CPAC-CNN models, increase the rank by 1 will incur a $0.003$ and $0.005$ seconds increase in the ``unit processing time", respectively. It is worth noting the above comparison is not completely fair and the reasons can be summarized into two aspects. Firstly, the original Conv layer is a built-in function in the Pytorch implemented by the NVIDIA CUDA Deep Neural Network library (cuDNN), which has already been specifically designed for acceleration while our proposed CPAC-Conv layer is implemented by the tensor multiplication methods without 
any computational acceleration design. Furthermore, the implementation of the CPAC-Conv layer can be improved by applying parallel computing, for example, the operations conducted by different kernel groups can run parallelly instead of sequentially, which can accelerate the CPAC-Conv layer computational implementation dynamically. In summary, the computational properties comparison in section 4.2.2 and Table 2 can show the large potential of using Tensor decomposition to compress deep learning.

{Additionally, we compared our proposed CPAC-Conv layer with Lebedev's work} (\cite{Leb15}). Considering an Original CNN model with one Conv layer and one FC layer, in our proposed CPAC-Conv model, the original CNN model will be approximated by one CPAC-Conv layer and one FC layer. In Lebedev's work, however, the original CNN will be approximated by four Conv layers with small kernels and one FC layer. Therefore, although Lebedev's method uses the same number of parameters as ours from the perspective of Conv layer parameters, their method increases the depth of the model by introducing additional layers. The increase of model depth will further introduce more intermediate variables which will take more memory. {By comparing our proposed CPAC-Conv layer and Lebedev's work, we can summarize that (1) our proposed CPAC-CNN (1 CPAC-Conv layer and 1 FC layer) receives quite similar performance with Lebedev's work (4 Conv layers with decomposed kernels and 1 FC layer) when selecting the same hyper-parameter; (2) to approximate the same CNN model (1 Conv layer and 1 FC layer), our proposed CPAC-CNN will use $4 MB$ less GPU memory compared with Lebedev's work (test on a single GTX 1080Ti GPU); (3) the training in Lebedev's method is a multi-step process, which includes a pre-train step and a finetune step. Our proposed CPAC-CNN can achieve single-step training to use less model learning time than Lebedev's work. From multiple metrics, i.e., accuracy, model complexity, required memory, and training time, we can conclude that our proposed method outperforms Lebedev's method.}

\subsection{Case 2: Magnetic Tile Defects Dataset}
\label{case2}
\begin{figure}[htp]
\centering
\resizebox*{14cm}{!}{\includegraphics{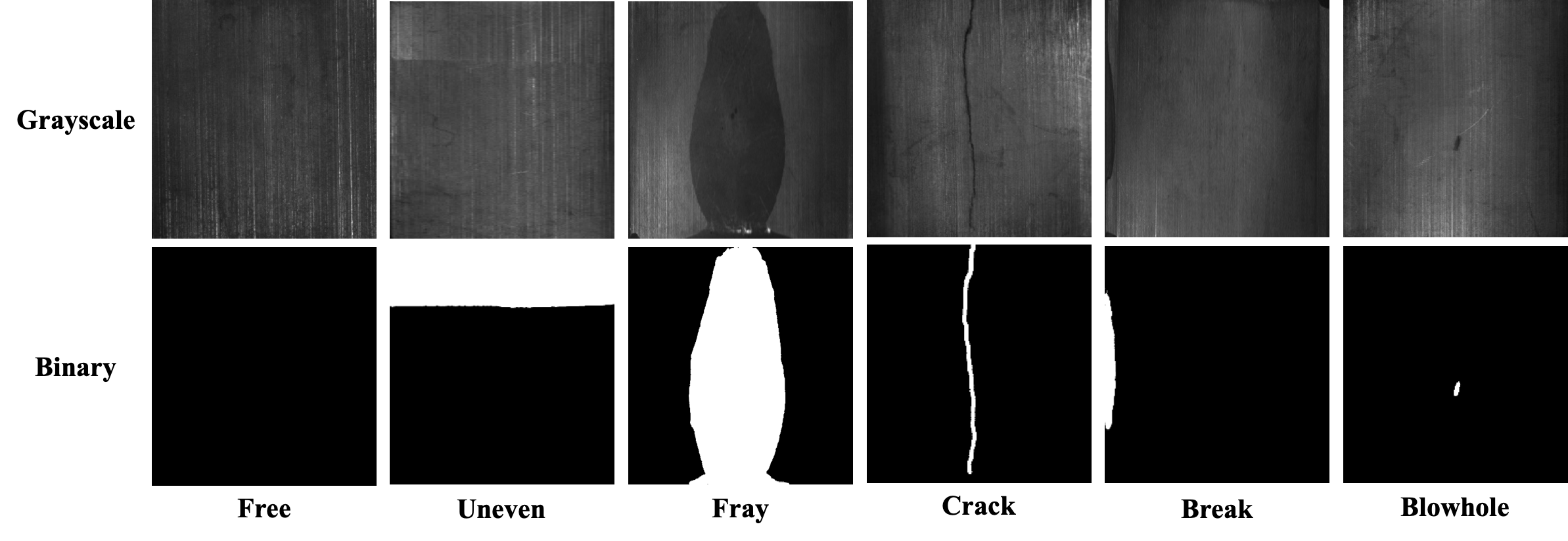}}
\caption{Image Data of Magnetic Tile Defect}
\label{example}
\end{figure}

The Magnetic Tile Defects data are collected for anomaly detection and diagnosis in the magnetic tile automation process. The samples are shown in Figure \ref{example}. {There are six classes in this dataset. Five classes of them are different types of defects (blowhole, crack, break, fray, uneven) and one of them is a defect-free class}. Because it is an unbalanced dataset and there are much fewer samples belonging to crack and fray than other classes, we apply the data augmentation technique, such as flip and rotate, to enrich the dataset. Furthermore, we resize the image data for consistency. To test the model performance on defect diagnosis, we use the dataset containing 563 $100\times 100$ binary samples for training and $100$ samples for testing. We test the performances of single-layer and double-layer CPAC-CNN and compare them with the corresponding CNN model. The detailed experiment settings and results are summarized in Table \ref{Defect_Results}. {From this table, we can see that the number of parameters in CPAC-CNN is controlled by the rank ($R$). The CPAC-CNN can achieve similar performance with about 20\% parameters comparing to the original CNN. That is to say, the CPAC-CNN can compress the parameter size of CNN by about 80\%.} 

\begin{figure}[htp]
\centering
\resizebox*{14cm}{!}{\includegraphics{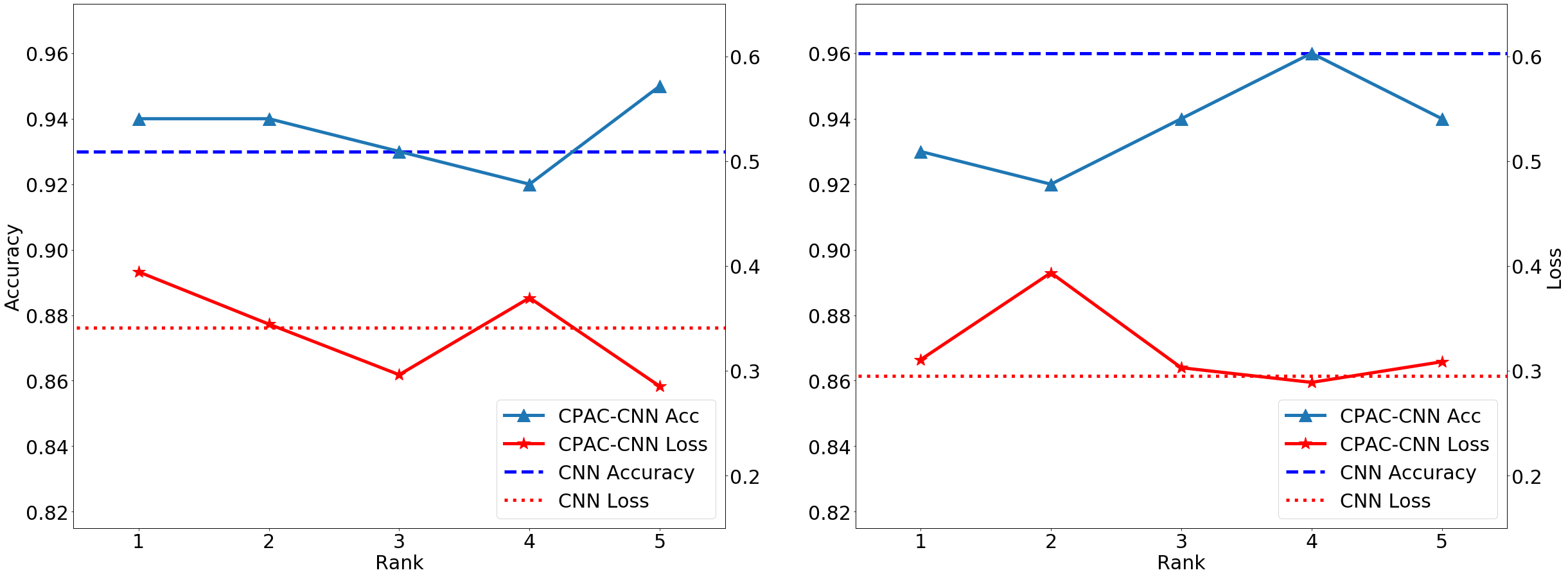}}
\caption{Performance Comparison Between CNN and CPAC-CNN on Magnetic Tile Defect, \textbf{Left:} one Conv/CPAC-Conv Layer, \textbf{Right:} two Conv/CPAC-Conv Layers}
\label{Defect_Results_Viz}
\end{figure}

\begin{table}[htp]
\newcommand{\tabincell}[2]{\begin{tabular}{@{}#1@{}}#2\end{tabular}}
\tbl{Experiment Results on Magnetic Tile Defects}
{\begin{tabular}{ccccccc} \toprule
 Model & Structure & Rank (R) & Kernel Size & \# of Parameters& CR\textsuperscript{*} & Accuracy \\
 \midrule
 CNN &  \tabincell{c}{1$\times$Conv Layer \\ 1$\times$FC Layer} & - & $(8,3,3,1)$ & 72 & 1 & 0.93\\
 \multirow{5}{*}{CPAC-CNN}&  \multirow{5}{*}{\tabincell{c}{1$\times$CPAC-Conv Layer \\ 1$\times$FC Layer}} & 1 & \multirow{5}{*}{$\tabincell{c}{$(8,R)$ \\ $(3,R)$ \\ $(3,R)$ \\ $(1,R)$}$} & \multirow{5}{*}{$15R$} & 0.2083 &0.94 \\
  &  & 2 &  &  & 0.4167 & 0.94\\
  &  & 3 &  &  & 0.6250 & 0.93\\
  &  & 4 &  &  & 0.8333 & 0.92\\
  &  & 5 &  &  & 1.0417 & 0.95\\
 \midrule
 CNN &  \tabincell{c}{2$\times$Conv Layer \\ 2$\times$FC Layer} & - & $(8,3,3,1),(8,3,3,8)$ & 648 & 1 & 0.96\\
  \multirow{5}{*}{CPAC-CNN}&  \multirow{5}{*}{\tabincell{c}{2$\times$CPAC-Conv Layer \\ 2$\times$FC Layer}} & 1 & \multirow{5}{*}{$\tabincell{c}{$(8,R),(8,R)$ \\ $(3,R),(3,R)$ \\ $(3,R),(3,R)$ \\ $(1,R),(8,R)$}$} & \multirow{5}{*}{$37R$} & 0.0571  & 0.93 \\
  &  & 2 &  &  & 0.1142  & 0.92\\
  &  & 3 &  &  & 0.1713  & 0.94\\
  &  & 4 &  &  & 0.2284  & 0.96\\
  &  & 5 &  &  & 0.2855  & 0.94\\
 \bottomrule
\end{tabular}}
\tabnote{\textsuperscript{*}CR: Compression Ratio}
\label{Defect_Results}
\end{table}
The visualization of performance comparison on defects diagnosis are shown in Figure \ref{Defect_Results_Viz}. Similarly, the left plot shows the comparison between single-layer models while the right plot shows the comparison between double-layer models. In Figure 6, we can see that the classification accuracy of CPAC-CNN (lines with triangle markers) is comparable to CNN (dashed lines), and the value of loss function in CPAC-CNN (lines with star markers) also approaches CNN (dotted lines). {In addition, we can figure out that the proposed CPAC-Conv layer can realize significant parameter compression without delaying the performance. Specifically, we can compress the number of parameters by about 80\% compared with conventional CNN to receive the similar testing accuracy.} 

Based on the relationship between the rank of CP-decomposition and testing accuracy shown in Figures \ref{MNIST_Results_Viz} and \ref{Defect_Results_Viz}, we can provide two principles in selecting the value of rank in using the CPAC-Conv layer. On the one hand, if the top priority is to maintain a comparable performance to the original Conv layer, the recommended compression ratio is around $20\%$ and the corresponding value of rank can be calculated accordingly. On the other hand, if the top priority is to compress the model as much as possible with acceptable performance, we would recommend starting from the rank $R=1$ and tuning the value of rank according to the accuracy expectation.

{We understand that over-parameterization could be one of the main reasons that the deep neural networks received outstanding performance} (\cite{DBLP:journals/corr/NeyshaburBMS17}). {But in engineering practice, according to the Occam's Razor principle, when presented with competing models about the same prediction, one should select the solution with the fewest parameters. Based on the experiment results in Tables} \ref{MNIST_Results} and \ref{Defect_Results}, {the original design of CNN has a large redundancy in parameters. Our proposed method can provide an alternative way to control the model complexity. It can reduce the parameter redundancy without decaying its performance. Besides, it has the flexibility of adjusting different model complexity by selecting a suitable compression ratio. A model with fewer parameters usually provides us with better generalization and a better insight to the relationship between model parameters and extracted features, which will be discussed in the following section.}
\subsection{Interpretation of Feature Map}
From the results shown in sections \ref{case1} and \ref{case2}, we can conclude that the rank of CP-decomposition determines the number of parameters in the CPAC-Conv layer and further influence the model performance. As we discussed before, the CPAC-Conv layer serves as a feature extractor in CPAC-CNN. In this section,  we will further illustrate how our proposed method influences the extracted features.

We use the single-layer CPAC-CNN as the example, which consists of one CPAC-Conv layer as the feature extractor and one FC layer as the classifier. According to equation (\ref{4-2-3}), we will decompose the original convolutional kernel $\mathcal{K}$ into $R$ groups of small kernels $K_{r}^{S},K_{r}^{X},K_{r}^{Y},K_{r}^{N}, ( r=1,...,R)$, and then sum up the features extracted by each kernel group as the output feature map for further classification. {For example, the features extracted by the $i_{th}$ kernel group is calculated by equation} (\ref{5-3-1}).
\begin{align}
    \Tilde{\mathcal{V}}_{i} = \left(\left(\left(\Tilde{\mathcal{U}} \times_{3} K_{i}^{S}\right) \times_{2} K_{i}^{Y} \right)\times_{1} K_{i}^{X}\right) \circ K_{i}^{N}
    \label{5-3-1}
\end{align}
{In equation} (\ref{5-3-1}), {$\Tilde{\mathcal{V}}_{i}$ represents the feature map extracted by $i_{th}$ kernel group.}
\begin{figure}[H]
\centering
\resizebox*{14cm}{!}{\includegraphics{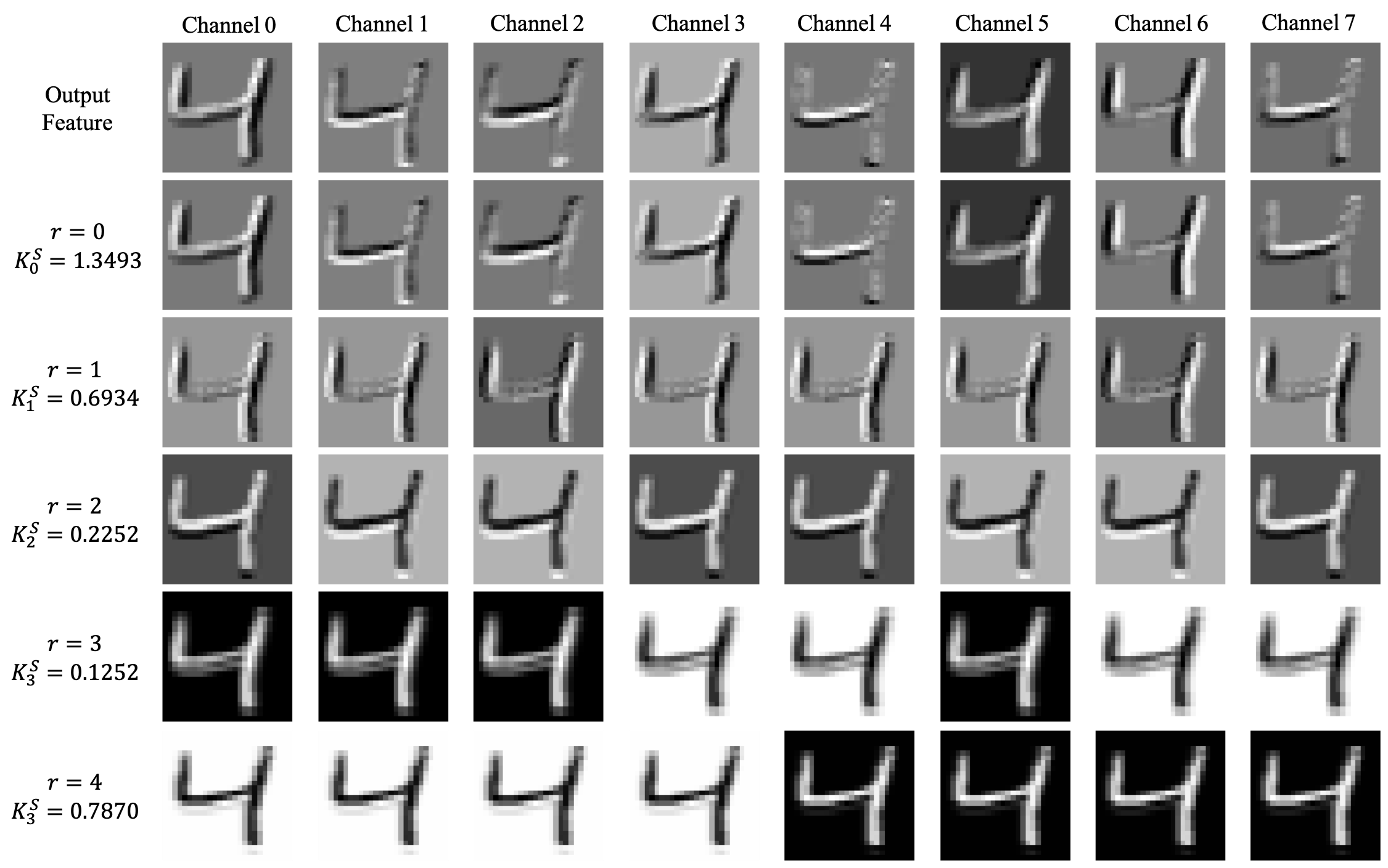}}
\caption{Relationship Between Features and Kernels on MNIST Data}
\label{MNIST_Interpret}
\end{figure}

\begin{figure}[ht]
\centering
\resizebox*{14cm}{!}{\includegraphics{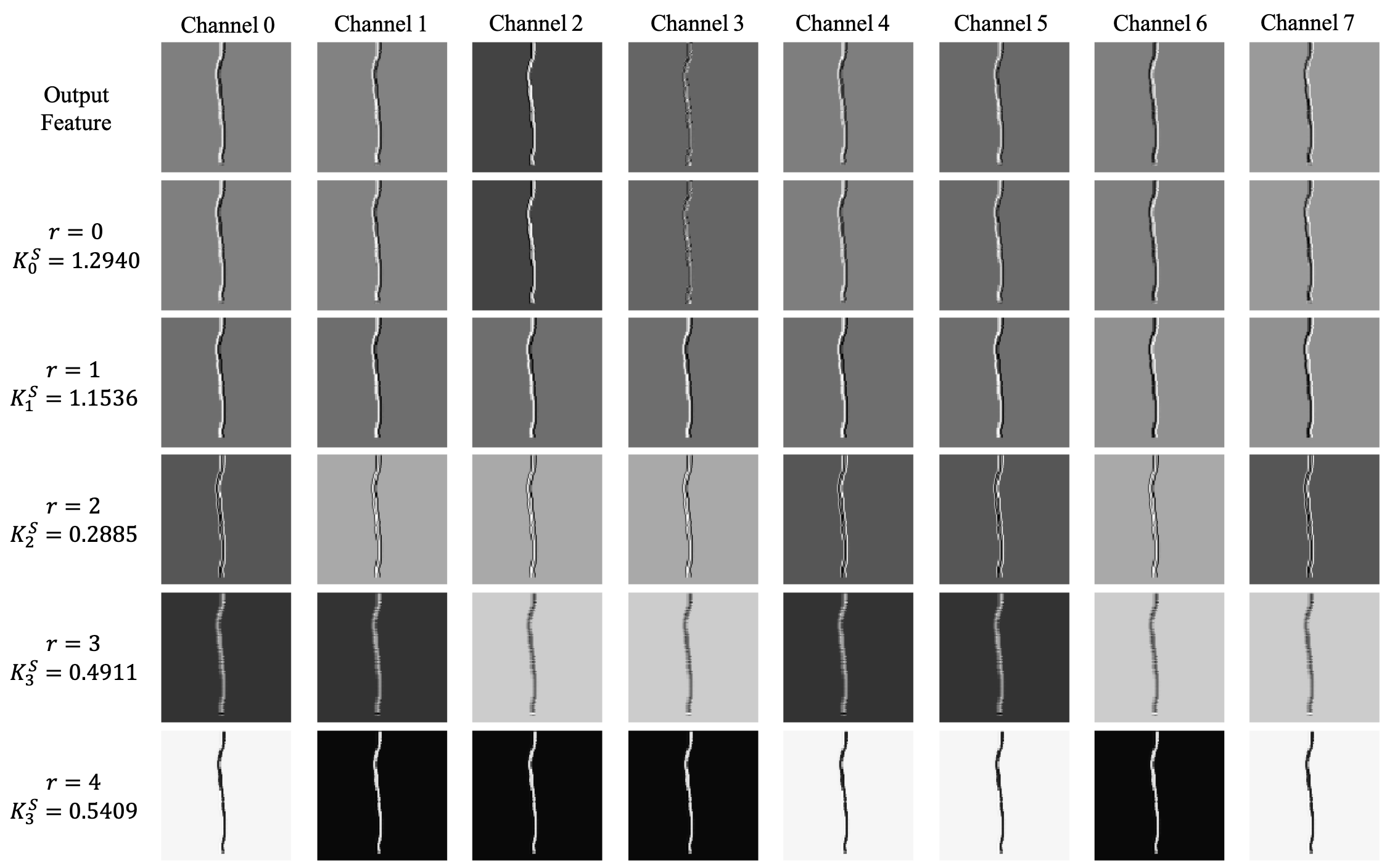}}
\caption{Relationship Between Features and Kernels on Defect Data}
\label{Defect_Interpret}
\end{figure}

To show the relationship between extracted features and kernels, we plot the overall feature map and features from each specific kernel group in Figures \ref{MNIST_Interpret} and \ref{Defect_Interpret}.

We plot the feature maps from single-layer CPAC-CNN and set the rank $R=5$ according to the results in Tables \ref{MNIST_Results} and \ref{Defect_Results} so that our proposed CPAC-CNN contains 5 kernel groups. There are 6 rows and 8 columns in both Figures \ref{MNIST_Interpret} and \ref{Defect_Interpret}. Each column represents one channel of the output feature maps, the first row represents the overall output feature maps, and each of the rest 5 rows represents features extracted from the corresponding kernel group. By analyzing these two figures, we can find out that the features extracted from the kernel group with the largest $K_{r}^{S}$ share almost the same pattern as the overall output feature. If we consider the kernel groups as a set of basis decomposed from the original convolutional kernel, one with the largest $K_{r}^{S}$ can be regarded as the most significant basis and output the most informative features. In this perspective, the value of $K_{r}^{S}$ could represent the significance of corresponding features, and if we use the CPAC-Conv layer as the feature extractor, we could select the most significant kernel group to get the most informative feature to further reduce the computation cost and improve model efficiency.

\section{Conclusion}
\label{conclude}
With the theoretical investigation and numerical analysis, we proposed the CPAC-Conv layer and further built the CPAC-CNN to compress the original CNN model without decaying the performance. In our proposed CPAC-Conv layer, we decompose the convolutional kernel into $R$ kernel groups and derive the mathematical expressions of its forward and backward propagations. With the help of CP-decomposition, we could approximate the original convolution operation with much fewer parameters, which reduces the model complexity and computation cost. And then, we propose the general setup of the CPAC-CNN model along with its training algorithm. In this way, we could stack multiple CPAC-Conv layers and use it with other types of layers to build neural networks for various tasks. Finally, as a feature extractor, we find out the value of $K_{r}^{S}$ indicates the significance of the feature map output from this kernel group, which could help us to interpret the importance of output features and provide us an indicator in feature selection.

{Given the properties of our proposed CPAC-Conv layer, we can conclude its contributions to industrial applications into three aspects. Firstly, fog computing is becoming increasingly used in manufacturing systems, in which the edge devices carry out a substantial amount of tasks but usually have limited computational resources. Compressing the model complexity will potentially reduce the computational requirements for deep neural network implementation in edge devices. Secondly, our proposed CPAC-Conv layer is an alternative to replace the original Conv layer. It is not limited to process image data. It can also be used to compress the deep neural networks with conv layers for processing videos and point clouds. Our proposed CPAC-Conv layer can be used as one alternative building block for new deep learning methods. Lastly, the model with fewer parameters is more promising according to Occam's Razor principle. It can have better interpretability and help us to understand the relationship between the decomposed kernels and extracted features.}

\section*{Acknowledgment}
This work was partially financially supported by the Department of Defense (DoD) MEEP program under award N00014-19-1-2728.
\bibliography{references}
\end{document}